\documentclass[10pt,twocolumn,letterpaper]{article}
\usepackage[dvipsnames]{xcolor}

\newcommand\blfootnote[1]{%
  \begingroup
  \renewcommand\thefootnote{}\footnote{#1}%
  \addtocounter{footnote}{-1}%
  \endgroup
}

\definecolor{cvprblue}{rgb}{0.21,0.49,0.74}
\usepackage[pagebackref,breaklinks,colorlinks,citecolor=cvprblue]{hyperref}

\usepackage{cancel}
\usepackage[accsupp]{axessibility}

\def\paperID{1978} % *** Enter the Paper ID here
\def\confName{CVPR}
\def\confYear{2024}

\title{Unleashing Unlabeled Data: A Paradigm for Cross-View Geo-Localization}

\author{Guopeng Li$^{1}$, Ming Qian$^{2}$, Gui-Song Xia$^{1,2,\dag}$\\
$^{1}$School of Computer Science, 
$^{2}$State Key Lab. LIESMARS, Wuhan University\\
{\tt\small \{guopengli, mingqian, guisong.xia\}@whu.edu.cn}
}

\usepackage{cvpr}              % To produce the CAMERA-READY version
\usepackage{graphicx}
\usepackage{amsmath}
\usepackage{amssymb}
\usepackage{array}
\usepackage{booktabs}
\usepackage[ruled]{algorithm2e}
\usepackage{algorithmic}
\usepackage{threeparttable}
\usepackage{multirow}
\usepackage{times}
\usepackage{epsfig}
\usepackage{graphicx}
\usepackage{setspace}
\usepackage{bbding}
\usepackage{stfloats}
\usepackage{colortbl}
\usepackage{bbding}

\usepackage[pagebackref,breaklinks,colorlinks,citecolor=cvprblue]{hyperref}

\usepackage[capitalize]{cleveref}
\crefname{section}{Sec.}{Secs.}
\Crefname{section}{Section}{Sections}
\Crefname{table}{Table}{Tables}
\crefname{table}{Tab.}{Tabs.}

\def\confName{CVPR}
\def\confYear{2024}

\begin{document}

\maketitle

\blfootnote{$^\dag$ Corresponding author}

%%%%%%%%% ABSTRACT
\begin{abstract}

This paper investigates the effective utilization of unlabeled data for large-area cross-view geo-localization (CVGL), encompassing both unsupervised and semi-supervised settings. Common approaches to CVGL rely on ground-satellite image pairs and employ label-driven supervised training. However, the cost of collecting precise cross-view image pairs hinders the deployment of CVGL in real-life scenarios. Without the pairs, CVGL will be more challenging to handle the significant imaging and spatial gaps between ground and satellite images. 
To this end, we propose an unsupervised framework including a cross-view projection to guide the model for retrieving initial pseudo-labels and a fast re-ranking mechanism to refine the pseudo-labels by leveraging the fact that ``the perfectly paired ground-satellite image is located in a unique and identical scene". The framework exhibits competitive performance compared with supervised works on three open-source benchmarks. Our code and models will be released on \url{https://github.com/liguopeng0923/UCVGL}.

\end{abstract}
%%%%%%%%% BODY TEXT

\vspace{-0.2cm}
\section{Introduction}
\label{sec:intro}

\begin{figure}[!t]
\centering
\includegraphics[width=\linewidth]{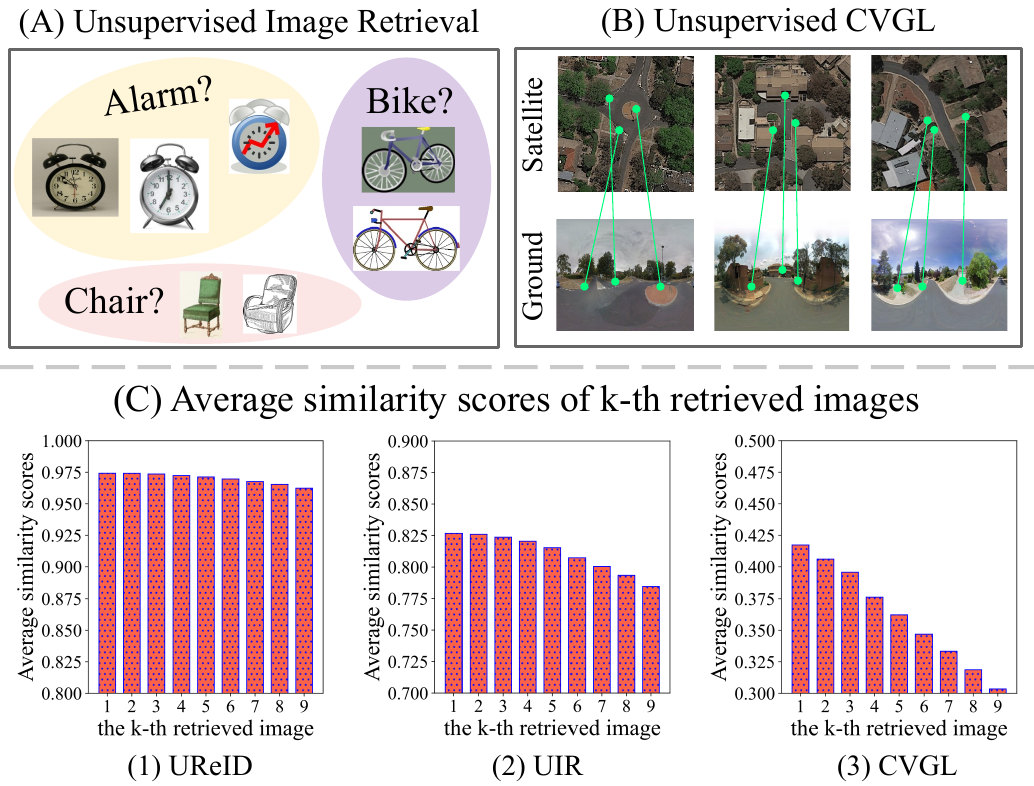}
  \caption{\textbf{Task Settings.} (A): Unsupervised Image Retrieval~\cite{hu2022feature} has object-level images and applies class-level clusters by class semantics. (B): Unsupervised CVGL has scene-level images and applies cross-view alignments by spatial correspondences (\ie, green lines). Compared with supervised CVGL, GPS labels and paired annotations (\ie, correspondences between ground and satellite images) are not accessible in UCVGL. (C): Common image retrievals are class-level (\eg, UReID~\cite{he2020fastreid} and UIR~\cite{hu2022feature}), but CVGL~\cite{Sample4Geo} aims to align cross-view images in the same scene, which is fine-grained instance-level~\cite{Sample4Geo}. The Top-k most similar images are more discriminative without class semantics in CVGL.}
  \label{fig:pic/1/different_image_retrival.pdf}
\vspace{-2mm}
\end{figure}

Large-area Cross-View Geo-Localization (CVGL) aims to determine the localization of ground images by retrieving the most similar GPS-tagged satellite images~\cite{zhu2022transgeo}. The easy availability of open-source GPS-tagged satellite images, such as Google Maps, helps users get accurate ground localization at a low cost ~\cite{Sample4Geo,wang2023finegrained} and provides CVGL with great potential for various real-life applications, including person localization, automatic navigation, and augmented reality ~\cite{Sample4Geo, li2019cross, mithun2023cross}.

However, existing CVGL methods are usually supervised training with ground-truth correspondences of Ground images (Grds) and Satellite images (Sats), known as SCVGL. This label-driven nature brings some practical limitations. Firstly, obtaining accurately located ground images requires expensive devices (\eg, lidar devices~\cite{geiger2013vision,huang2024sunshine}), and matching ground and satellite images brings extra human costs~\cite {VIGOR}. Secondly, numerous non-corresponding ground and satellite images without GPS tags are available online but cannot be directly used for SCVGL. Lastly, SCVGL requires re-annotation for new or changed scenes, increasing the cost of human effort and resources. Given these limitations, an important question arises: can we uncover hidden correlations from cross-view images without label information? This motivates us to study how to utilize unlabeled data to start a retrieval system for CVGL.

Regrettably, just as a ship needs a compass to find its way, without labels, the model may lose its optimized direction. This phenomenon is usually called \emph{cold-start problem}~\cite{mannix2023cold} for un- and semi-supervised learning.

To mitigate this issue, most unsupervised image retrievals (UIRs) label an informative subset of images first, guided by some common prior knowledge (\eg, category~\cite{hu2022feature} or instance~\cite{he2020fastreid, chen2022deep} information). Compared to them, CVGL has some unique characteristics illustrated in \cref{fig:pic/1/different_image_retrival.pdf} and more details in \cref{sec: Unsupervised Image Retrieval (UIR)}. Specifically, existing retrievals ~\cite{hu2022feature,he2020fastreid,deng2022insclr, chen2022deep} have object-level cluster centers, such as some alarm, person or object ID, which includes many images from the same category or instance and enlighten them to apply cluster technologies ~\cite{dbscan,kmeans} to get initial label sets, as can be seen in the upper row of \cref{fig:pic/1/different_image_retrival.pdf}. Differently, most CVGL~\cite{Sample4Geo,zhu2022transgeo} methods are trained by one-to-one cross-view image pairs to learn spatial correspondences, which means, ``class" information is not essential in the training process and CVGL is a fine-grained retrieval. The reference satellite images can only be retrieved by ground images in the same scene position by spatial correspondences, just like green lines in \cref{fig:pic/1/different_image_retrival.pdf}(B). Besides, in \cref{fig:pic/1/different_image_retrival.pdf}(C), we collect the similarity scores between each image and its $k$-th retrieved image and average the scores of more than 5000 image pairs in different tasks. It's clear that compared to existing instance retrieval ~\cite{he2020fastreid} and image retrieval ~\cite{hu2022feature}, the average similarity scores of CVGL~\cite{Sample4Geo} are lower and drop fast from 1-st to the right. This comparison further suggests our task is non-class and fine-grained. Therefore, we cannot cold-start the unsupervised CVGL using existing image retrieval methods.

In this paper, we design a framework for unsupervised and semi-supervised CVGL. Following a common design ~\cite{cho2022part,hu2022feature}, in the unsupervised setting, we separate our framework into two components: the cold-start stage to produce initial pseudo-labels and the semi-supervised stage to refine labels. Moreover, the semi-supervised stage can also be utilized alone for CVGL (details in \cref{sec:semi-supervised-results}). In our framework, non-class and spatial corresponding characteristics are fully utilized, making the unsupervised CVGL possible and the semi-supervised CVGL a good performance.

In the cold-start stage, the significant imaging and view gaps make ground and satellite images distinct features\cite{safa} in the same scene. To address this issue, we attract cross-view images in the same position by spatial corresponding prior knowledge. Specifically, we project the ground panorama into a 3D coordinate system and obtain its bird's-eye-view image through a spherical transform ~\cite{shi2022accurate,wang2023finegrained}. We then fill the unknown regions and reduce imaging gaps by a CycleGAN model ~\cite{CycleGAN2017}, resulting in a fake satellite image (\eg, \cref{overall.pdf}(B)) aligned with the ground panorama. Finally, we replace unknown ground-satellite pairs with known ground-fake pairs as trained positive pairs, enabling successful retrieval of more than 40\% image pairs on CVACT and achieving cold-start initialization.

In the semi-supervised stage, the initial labels may be noisy or insufficient, which limits the final performance. We re-rank pseudo-labels to enhance the quality and quantity of pseudo-labels by using non-class characteristics. Specifically, the perfectly paired cross-view images are two views of the same scene, which should be mutually retrievable and dissimilar to other images, as shown in \cref{fig:pic/1/different_image_retrival.pdf}(C-3). We filter out many ``error" labels that do not meet this requirement with a threshold, improving the correctness ratios of the remaining labels from around 30\% to 80\% on CVACT. However, the number of pseudo-labels is only about 20\% of the dataset. Therefore, as the model's knowledge increases, we progressively introduce more challenging samples by adjusting our filtering mechanisms to get more pseudo-labels. Eventually, our model achieves competitive performance with recent supervised approaches ~\cite{zhu2022transgeo,zhu2023simple,Sample4Geo}.

\section{Releated Work}
%-------------------------------------------------------------------------

\subsection{Large-area Cross-View Geo-Localization}

Ground images (Grds) and Satellite images (Sats) are usually captured by different sensors with huge perspective differences and different environmental conditions, making CVGL challenging.

In order to simplify CVGL, many works leverage {two assumptions\label{two assumptions}} in popular datasets~\cite{CVUSA, CVACT, VIGOR}. 1) Localization Alignment~\cite{zhu2023simple}: the localization of Grds is aligned to a known position of Sats. 2) Orientation Alignment~\cite{shi2020looking}: the geographical orientation of Grds is aligned to Sats. By leveraging these alignments, some works ~\cite{vo2016localizing,cai2019ground,liu2019lending,regmi2019bridging,shi2020optimal} implicitly enhance the high-level features supervised by ground-truth correspondences. Recently, some methods~\cite{safa,Toker_2021_CVPR,zhang2023cross,wang2021each} improve the spatial alignments of input images by the explicit polar transform for better performance, but the transformed images suffer from severe distortions that also need to be corrected by full supervisions.

To make CVGL more practical, some methods try to get rid of one of two assumptions. Firstly, ~\cite{safa,zhu2022transgeo,yang2021cross,zhu2023simple,Sample4Geo} have achieved good results without relying on Localization Alignment, but they still rely highly on ground-truth correspondences or GPS labels~\cite{Sample4Geo}.  Secondly, although some works~\cite{hu2018cvm,zhu2021revisiting,shi2020looking,mithun2023cross} make efforts to remove the Orientation Alignment, they need to align the rotation angle of ground images and polar transformed images ~\cite{safa,shi2020looking} with known cross-view image pairs~\cite{zhu2022transgeo}. 

Different from existing works, in this paper, we no longer rely on ground-truth labels and instead propose a more challenging and practical setting called Unsupervised CVGL (UCVGL) to unleash unlabeled data.

\subsection{Unsupervised Image Retrieval (UIR) \label{sec: Unsupervised Image Retrieval (UIR)}}
Unsupervised Image Retrieval (UIR) is a well-studied area, but Unsupervised Cross-View Geo-Localization (UCVGL) presents unique challenges and characteristics that require specialized approaches. Existing image retrieval methods are not directly applicable to UCVGL due to the specific requirements of fine-grained, cross-domain, and cross-view retrieval ~\cite{zhang2022dataset,deng2022insclr,Hu_2023_ICCV,hu2022feature,chen2020mocov2,chen2021ice,sarlin2020superglue}.

Firstly, traditional instance-level methods aim to retrieve all reference images containing the object depicted in the query image~\cite{he2020fastreid,hu2022feature}, generating pseudo labels through clustering similar features. However, this approach is not suitable for UCVGL, which necessitates fine-grained retrieval based on unique spatial correspondences ~\cite{zhu2022transgeo}.

Secondly, Unsupervised Cross-Domain Image Retrieval~\cite{hu2022feature} aligns two different domains by reducing the distances between limited class centroids using self-supervised pre-trained models ~\cite{Hu_2023_ICCV,hu2022feature,chen2020mocov2,zhang2022tree,wang2023contrastive}. However, aligning cross-view images in UCVGL is challenging due to the absence of class semantics and large spatial gaps. Besides, self-supervised pre-trained models are unnecessary for UCVGL.

Lastly, existing cross-view retrieval methods often focus on handling minor view variances through feature matching technologies ~\cite{sarlin2020superglue} or cluster algorithms~\cite{chen2021ice,deng2022insclr}. However, these methods are inadequate for CVGL as ground and satellite images have significant perspective differences, occlusions, and illumination variations.

In summary, the objectives and settings of existing retrievals do not align with the requirements of UCVGL. This necessitates the development of specialized approaches that consider the unique challenges posed by fine-grained, cross-domain, and cross-view retrieval in UCVGL.

\begin{figure*}[!t]
  \centering
  \centering
    \includegraphics[width=\linewidth]{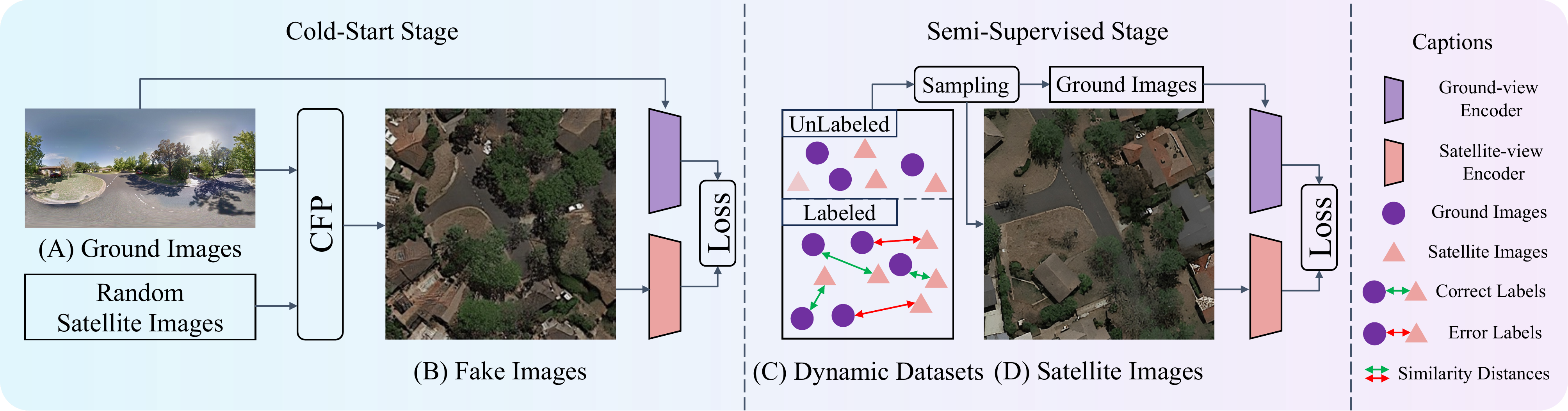}
    \caption{\textbf{Pipeline Overview.} Firstly, we train two separate encoders with ground panoramas and projected images to initialize a cross-view feature space for solving cold-start problems. Secondly, we train ground-satellite image pairs by sampling from adaptive pseudo-labels.}
  \label{overall.pdf}
\vspace{-0.2cm}

\end{figure*}

\section{Method}

\textbf{Problem Statement:} Given the set of ground images $\{\rm{I_{grd}}\}$ and satellite images $\{\rm{I_{sat}}\}$, the objective of CVGL is to learn an embedding space where each $\rm{I_{grd}}$ is close to its \emph{nearest} corresponding ground-truth $\rm{I_{sat}}$~\cite{zhu2022transgeo}. In SCVGL~\cite{Sample4Geo}, each ground-view image and its corresponding satellite image are trained as a positive pair, and other cross-view pairs are trained as negative. Differently, we propose a more practical setting, \ie Unsupervised CVGL (UCVGL), which trains the models without GPS labels and the ground-truth correspondences between ground and satellite images.

\noindent \textbf{Method Overview:} Our framework can implement cold-start initialization for unsupervised learning and separate semi-supervised learning in CVGL, as illustrated in \cref{overall.pdf}. Without any labels, we design a cold-start stage to align cross-view images and label some data as pseudo-labels. Subsequently, we refine this pseudo-label set or a given label set, and 
inject more samples adaptively as the models' knowledge increases.

\noindent \textbf{Soft Symmetrical InfoNCE Loss:}
Our encoders are trained with Symmetric InfoNCE Loss ~\cite{he2020momentum, Sample4Geo}:
\begin{equation}
\mathcal{L}(q, R)_{\mathrm{InfoNCE}} = -\log\left(\frac{\exp(q \cdot r_{+} / \tau)}{\sum_{i=0}^{R}\exp(q \cdot r_{i}/\tau)}\right)
\label{eq:symmetric_infonce_loss}
\end{equation}
Here, $q$ denotes an encoded query image, and $R$ is a set of encoded reference images. The InfoNCE loss is low when the query $q$ and the positive match $r_{+}$ are similar and high when they are dissimilar. The temperature parameter $\tau$ is learnable. Given the prevalence of noisy labels throughout the training process, such as fake images in (B) and error labels in (C) of \cref{overall.pdf}, we additionally smooth the loss using label smoothing ~\cite{szegedy2016rethinking}.

\subsection{Cold-Start for unsupervised learning}
Supervised methods~\cite{zhu2023simple,Sample4Geo} enable robust retrievals by human-annotated labels, but unsupervised approaches struggle to understand what retrieval feature spaces users want without labels. This phenomenon, known as the \emph{cold-start problem}~\cite{mannix2023cold,Zheng_2022_CVPR}, poses one of the most challenging obstacles in unsupervised learning for UCVGL. It implies the absence of direct information to align cross-view images in the same position with unlabeled ground-satellite databases. To this end, we design a Correspondence-Free Projection to project ground panoramas to satellite view by leveraging the spatial correspondences of cross-view images and imaging principle. The transformed images, such as (B) in \cref{overall.pdf}, have the same position as ground images and resemble satellite images ((D) in \cref{overall.pdf}). Then, we cleverly design a self-supervised objective to attract ground and projected images, leading to robust cross-view feature alignments. Lastly, our model successfully predicts more than 40\% cross-view image pairs and provides 6012 pseudo labels with about 80\% correct ratios (CVACT in \cref{table:stage-0}).
\begin{figure}[!t]
\centering
\includegraphics[width=\linewidth]{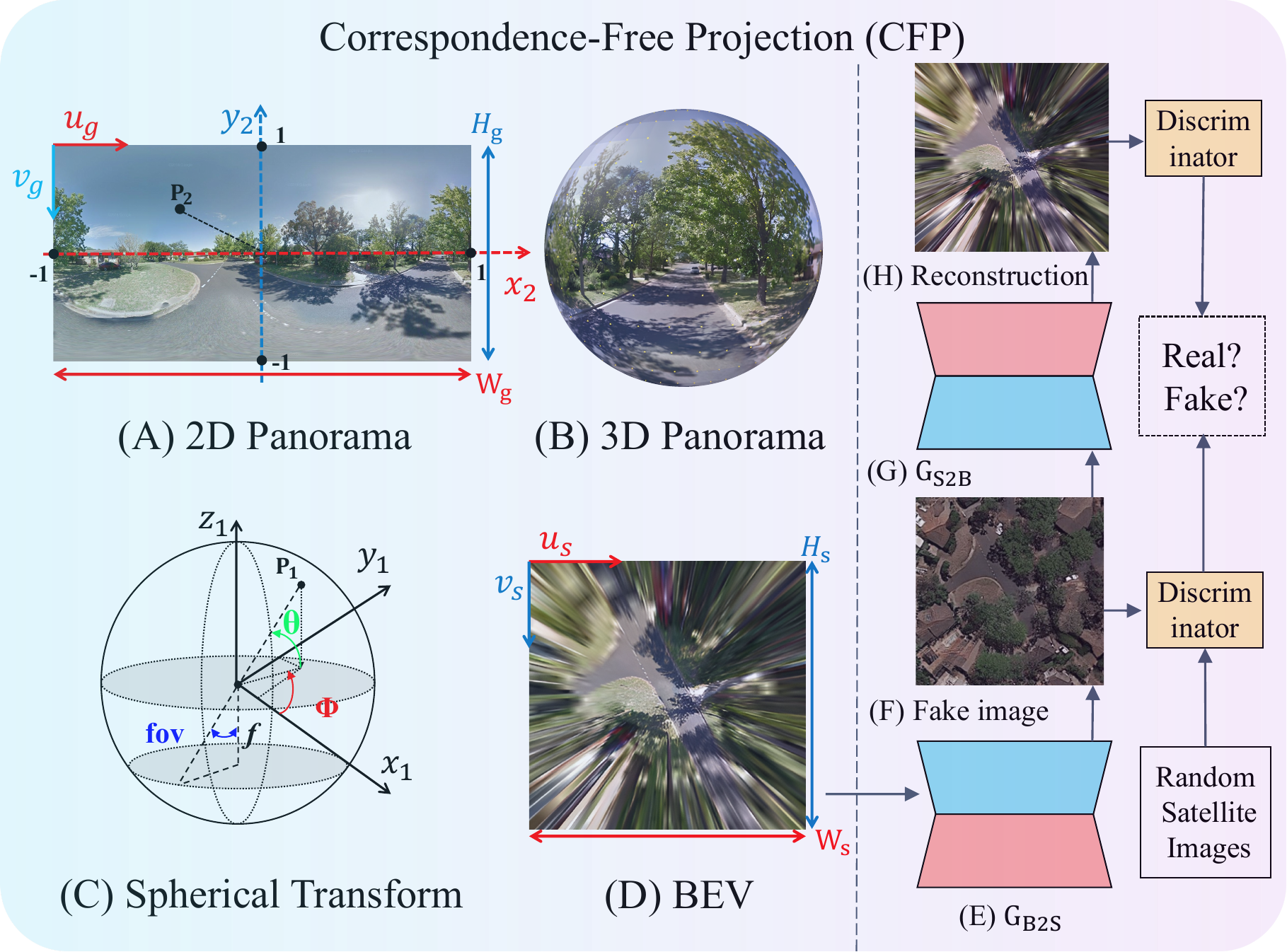}
  \caption{\textbf{Projections.} We project geometrically ground panoramas to BEV-view images on the left and transform BEV into fake images that resemble satellite images on the right.}
  \label{fig:pic/3/projection.pdf}
  \vspace{-0.5cm}
\end{figure}
\vspace{-0.1cm}
\subsubsection{Correspondence-Free Projection (CFP)\label{sec:CFP}}
Without correspondences between  $\{\rm{I_{grd}}\}$ and $\{\rm{I_{sat}}\}$, the implicit spatial alignments are hard to learn which forces us to leverage unique data-driven knowledge. We observe two differences in cross-view images: (a) they have very different perspectives because they are captured in the front-view and top-view camera positions. (b) they have different styles of images caused by different camera sensors and imaging processes. In this section, we bridge the two gaps by Geometric Projecting and Imaging Projecting without any ground-truth correspondences, and the projected images achieve explicit spatial alignments from ground-view images to satellite-view images.

\noindent \textbf{Geometric Projection} ~\cite{shi2022accurate,wang2023finegrained}: 
As shown in (A-D) of \cref{fig:pic/3/projection.pdf}, we denote the points of the 3D world by $\mathrm{P}_{1} = (x_{1}, y_{1}, z_{1})$ or $\mathrm{P}_{1} = (\phi,\theta)$, the field of view of the bird's-eye view (BEV) by $fov$, the points of panoramas (the width and height are $\mathrm{W}_g$ and $\mathrm{H}_g$) by $\mathrm{P}_{2} = (u_{g}, v_{g})$, and the points of BEV-images (the width and height are $\mathrm{W}_s$ and $\mathrm{H}_s$) by $\mathrm{P}_{3} = (u_{s},v_{s})$. The focal length of the BEV in the imaging process is $f = 0.5\mathrm{W}_s / tan(fov)$. We can transform the points $\mathrm{P}_{2}$ to $\mathrm{P}_{3}$ according to the captured principles in the 3D spherical world:
\begin{equation}
\label{geometric Transform}
\left\{
\begin{aligned}
u_g = [1 - &{\rm arctan2}(\mathrm{W}_s/2-u_s,\mathrm{H}_{s}/2-v_s)/\pi]\mathrm{W}_{g}/2   \\  
&v_g = [0.5-{\rm arctan2}(-f,d/\pi)]\mathrm{H}_g
\end{aligned}
\right.
\end{equation}
where $d=\sqrt{(\mathrm{W}_s/2-u_s)^{2}+(\mathrm{H}_s/2-v_s)^{2}}$ denotes the distance between points of BEV and the camera's projective center. After the Geometric Projection, we obtained BEV images $\{\mathrm{I}_{\rm{bev}}\}$ that exhibit high geometric similarity to satellite images (\eg, (d) in \cref{fig:pic/3/projection.pdf}). 

\noindent \textbf{Imaging Projection}~\cite{CycleGAN2017,goodfellow2014generative}: While we get BEV images that resemble satellite images in geometry, these two kinds of images still have some imaging gaps such as different distortions, illumination, occlusion, weather, and so on. To this end, we apply a generative CycleGAN model~\cite{CycleGAN2017, kim2019u,ma2023humannerfse,zhao2023loco} to decrease further the imaging gaps between the two-view images illustrated in (D-H) of \cref{fig:pic/3/projection.pdf}. Specifically, we train the model with unpaired BEV and satellite images, after the training, the model can fill the distorted unknown areas in BEV and project the style of BEV to satellite-view images. By now, we obtain projected fake images $\{\mathrm{I}_{\rm{fake}}\}$ that exhibit high similarity to the remote sensing images (\eg, \cref{overall.pdf} (B)), suggesting the explicit projection aligns the spatial and imaging difference of cross-view images. Note that our projection is differentiable, correspondence-free, and performs one-to-one mappings from ground to fake images.

\begin{figure}[!t]
\centering
\includegraphics[width=\linewidth]{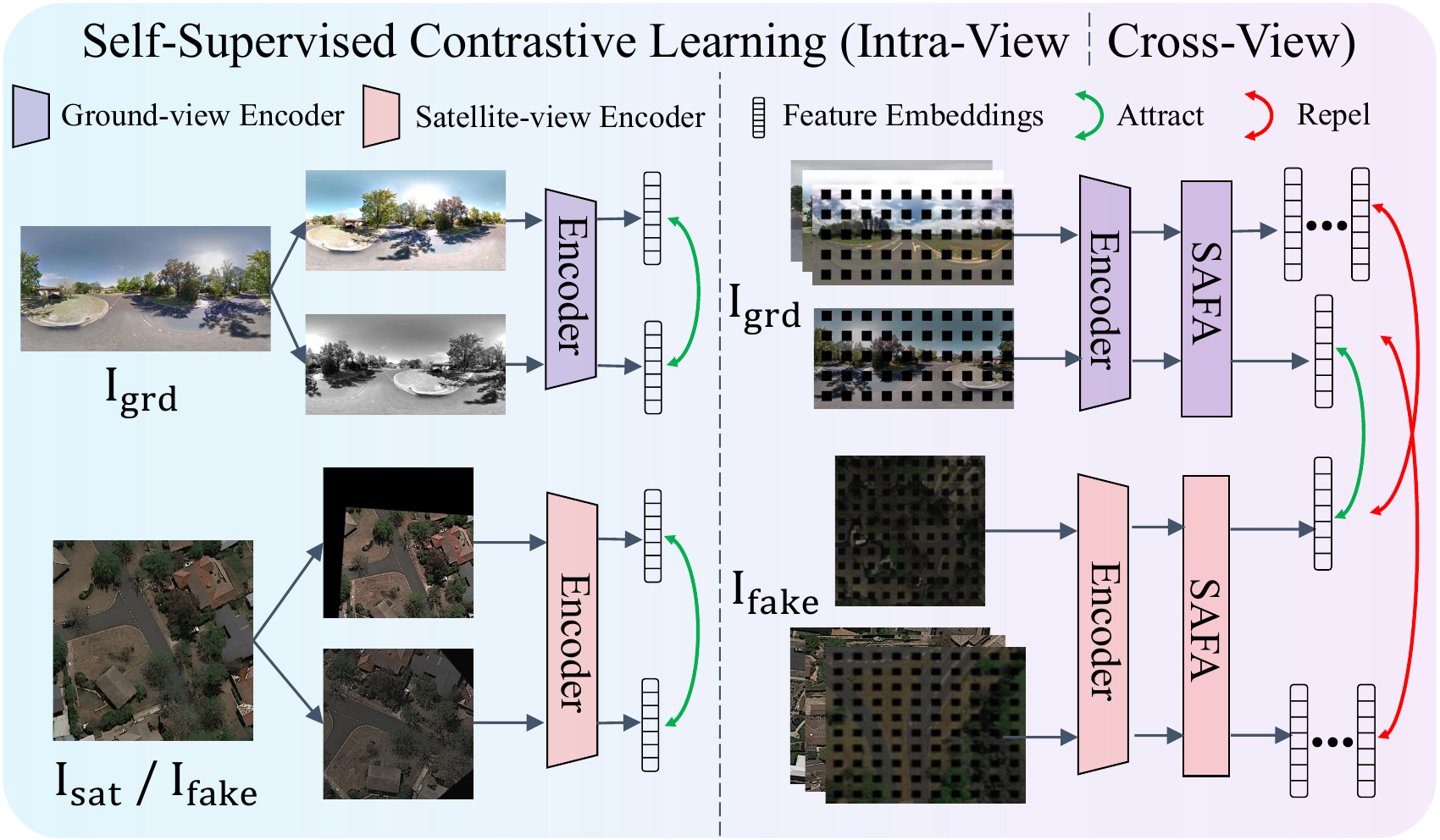}
  \caption{\textbf{Self-supervised contrastive learning.} We learn intra-view discriminative features by attracting two self-augmented images and cross-view alignments by attracting cleverly ground images and projected fake images.}
  \label{fig:self-supervised.pdf}

\end{figure}
\subsubsection{Self-Supervised Contrastive Learning\label{sec:self-supervised}}
Supervised CVGL methods~\cite{Sample4Geo} learn both discriminative instance-level and aligned spatial features by ground-satellite pairs. In UCVGL, the ground-satellite pairs are not accessible, but we can replace them with the previous ground-fake pairs produced by CFP. Specifically, we apply two self-supervised contrastive learnings illustrated in \cref{fig:self-supervised.pdf}: intra-view learning to make each scene more discriminative and cross-view learning to align the spatial features of cross-view images in the same scene position. 

\noindent \textbf{Intra-View:} Each same-view image is captured in an approximately independent scene in CVGL~\cite{Sample4Geo}. For this objective, we apply instance-wise contrastive learning methods that take only augmented images of the same image as positive pairs (\ie, $q$ and $r_{+}$ in \cref{eq:symmetric_infonce_loss}), while all other same-view samples in the dataset are regarded as negatives like~\cite{chen2020simple}. 
After intra-view learning, two encoders have an initial feature space to distinguish different scenes, which facilitates the convergence of after steps and benefits for the quality of pseudo-labels (details in \cref{table:stage-0}).

\noindent \textbf{Cross-View:} In CVGL~\cite{Sample4Geo}, each cross-view image pair is aligned because they approximately represent the same scene, such as ground-satellite image pairs. Although the ground-satellite image pairs are not accessible, our ground-fake image pairs generated by CFP are actually cross-view representations for the same scene.  So we think the model can learn robust spatial features to align the two views by training on ground-fake pairs. Specifically, a ground image and the corresponding fake image are a positive pair (\ie, $q$ and $r_{+}$ in \cref{eq:symmetric_infonce_loss}) and this ground image and other fake images are negative pairs. Advanced SCVGL methods (~\cite{Sample4Geo,zhu2022transgeo}) learn spatial features without any spatial enhancement module~\cite{zhu2023simple,safa,shi2020looking} by aligning ground-truth pairs, but it is hard for our UCVGL because of the noisy pseudo-labels and unavoidable inconsistency between fake and satellite images. Therefore, an additional spatial attention module SAFA~\cite{safa} is incorporated after the encoders to enhance spatial robustness.

\begin{figure}[!t]
\centering
\includegraphics[width=\linewidth]{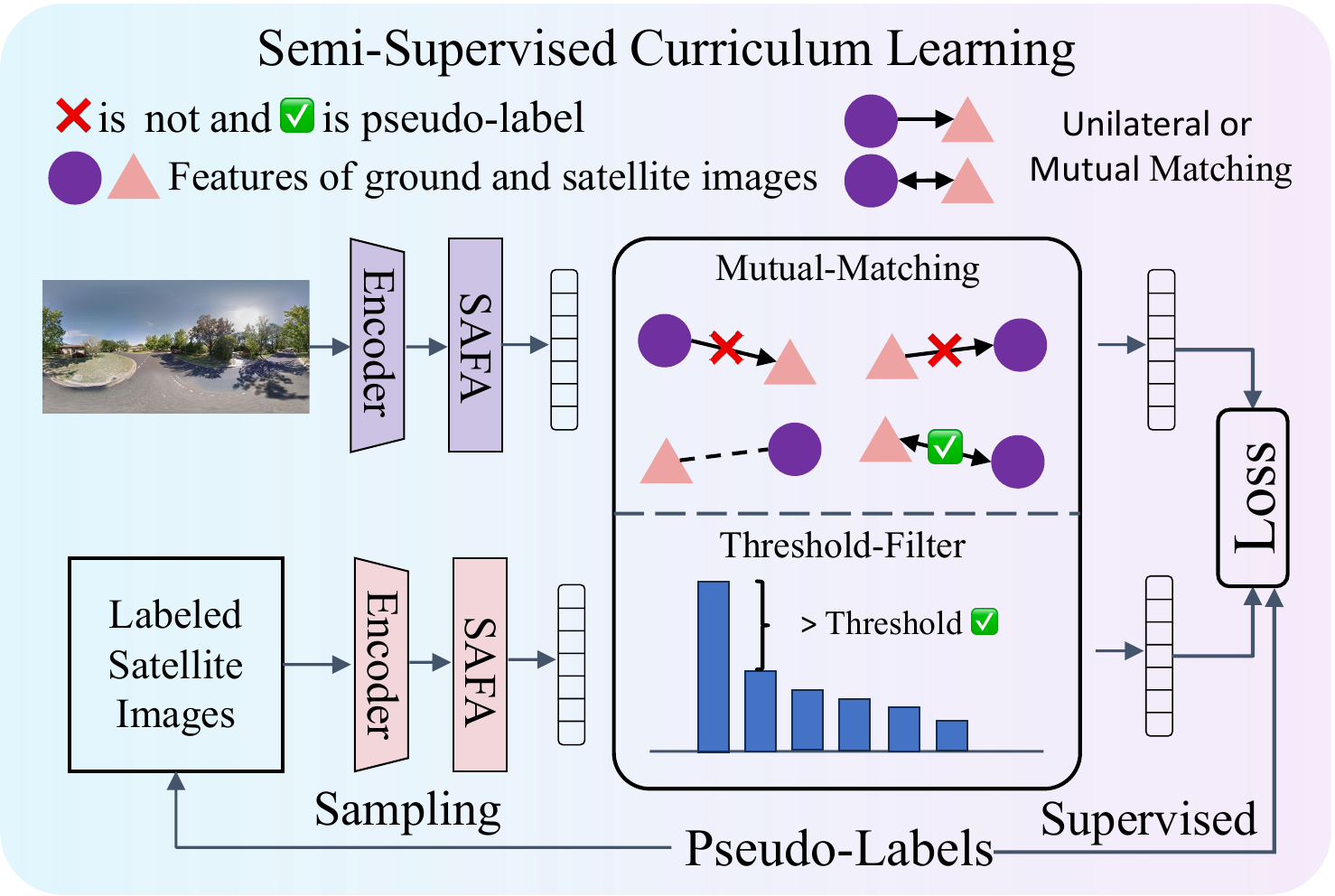}
  \caption{\textbf{Semi-supervised curriculum learning.} Ground images and satellite images are attracted and supervised through the guidance of adaptive pseudo-labels.}
  \vspace{-0.5cm}
  \label{semi-supervised.pdf}

\end{figure}
\subsection{Semi-Supervised Curriculum Learning\label{sec:semi-supervised}}
While the model trained by self-supervised contrastive learning aligns the two views to some extent and successfully predicts certain ground-truth correspondences between $\{\mathrm{I}_{\rm{grd}}\}$ and $\{\mathrm{I}_{\rm{sat}}\}$, the predictions exhibit excessive noise, rendering them unsuitable as pseudo-labels for direct supervisions (only about 30\% predictions are correct on CVACT in \cref{fig:pseudo-labels.pdf} (C-a)). In order to enhance the quality of pseudo labels, we propose a rapid matching mechanism termed Adaptive-Mutual-Matching (AMM), which selectively identifies ``true" correspondences from the predictions as pseudo-correspondences with a threshold filter. Based on these pseudo-labels (about 80\% labels are correct), we start semi-supervised learning to align the features of ground and satellite images.

\noindent \textbf{Mutual-Matching:} Due to the ground and satellite images are actually one scene with different views,  the
truly aligned ground-satellite correspondences should be retrievable mutually. Formulaly, given the extracted features of $i$-th ground image $\mathrm{I}_{\rm{grd}}^{~i}$ and the $j$-th satellite image $\mathrm{I}_{\rm{sat}}^{~j}$, they are considered as a truly positive pair (\ie, $q$ and $r_{+}$ in \cref{eq:symmetric_infonce_loss}) if and only if $\mathrm{I}_{\rm{grd}}^{~i}$ and $\mathrm{I}_{\rm{sat}}^{~j}$ are mutually the most similar to each other. In this paper, we use the cos similarity to evaluate the similarity like ~\cite{zhu2022transgeo}.  

\noindent\textbf{Threshold-Filter:} Although we filter most error labels by mutual matching, current pseudo-labels are also slightly noisy, especially for CVUSA (about 25\% correct ratio of labels after \emph{mutual-matching} ). In CVGL, the objective is to find the most similar reference image to the query image~\cite{zhu2022transgeo}, so the retrieved image pairs should have high similarity scores and be dissimilar from other images like (C-3) in \cref{fig:pic/1/different_image_retrival.pdf}. Therefore, we refine them using a threshold criterion. Specifically, we retain pseudo-pairs where the difference between the first-largest similarity and the second-largest similarity exceeds a threshold. This allows us to establish high-quality pseudo labels (more details in \cref{fig:pseudo-labels.pdf} (B)). However, the filtered labeled image pairs are often considered easy for models, suggesting a potential risk of overfitting if we solely rely on training with these high-quality pseudo-labels.

\noindent\textbf{Curriculum-Learning:} When humans learn knowledge, they often learn easy courses first and then difficult courses after they have a certain foundation~\cite{bengio2009curriculum}. Motivated by this, we train a ``stupid" model on easy samples, which are obtained by filtering the pseudo-labels using a high threshold.
As the model's knowledge advances, we introduce progressively more difficult samples by lowering the filtering threshold and establishing more mutual-matching pairs.
This iterative process ensures that our training samples diversify over time, mitigating the risk of overfitting. Besides, we also introduce other unlabeled data to learn negative pairs. Finally, we get a surprising performance that is comparable with the nearest supervised approach~\cite{Sample4Geo}.

\section{Experiment}
\subsection{Datasets and Metrics}
\textbf{Datasets.} We conducted experiments on three 
cross-view panorama datasets, \ie, CVUSA ~\cite{CVUSA,CVUSA2}, CVACT ~\cite{CVACT}, and VIGOR ~\cite{VIGOR}. CVUSA and CVACT consist of various images captured in rural and urban areas, each with 35,532 ground-satellite pairs for training and 8,884 pairs for evaluation. Besides, CVACT has an additional test dataset including 92802 image pairs. In the two datasets, each satellite-view image has one corresponding street-view image, and all pairs are aligned with the similar spatial localization~\cite{zhu2023simple} and orientation~\cite{shi2020looking}. We evaluate the results of our un- and semi-supervised settings in the two datasets. VIGOR has 105,214 panoramas and 90,618 aerial images from four cities. Due to the complex city scenes, the projection between ground and satellite images is hard on VIGOR, even in supervised settings~\cite{qian2023sat2density}. So it is difficult for us to solve the cold-start problem by simple homography projection~\cite{shi2022beyond} in unsupervised settings.  We provide a semi-supervised method for it. To the best of our knowledge, we are the first to remove GPS labels and ground-satellite pair annotations in CVGL.

\noindent\textbf{Evaluation Metrics.} We evaluate the retrieval performance by top-$k$ recall accuracy (R@$k$)~\cite{Sample4Geo}. Specifically, for each query ground image, we retrieve the $k$ nearest reference neighbors in the embedding spaces based on cosine similarity. A retrieval is considered correct if the ground-truth satellite image appears among the top $k$ retrieved images.
\begin{table*}[!htbp]
\resizebox{\textwidth}{!}{

\begin{tabular}{cc|cccc|cccc|cccc}
\hline \hline
\multirow{2}{*}{Approach} & \multirow{2}{*}{GT ratio} & \multicolumn{4}{c|}{CVUSA $\uparrow$}    & \multicolumn{4}{c|}{CVACT Val $\uparrow$} & \multicolumn{4}{c}{CVACT Test $\uparrow$} \\
                          &                       & R@1   & R@5   & R@10  & R@1\% & R@1    & R@5    & R@10  & R@1\% & R@1    & R@5    & R@10  & R@1\% \\ \hline
 \rowcolor{gray!20} CVM-NET~\cite{hu2018cvm}                   & 100\%            & 22.47 & 49.98 & 63.18 & 93.62 & 20.15  & 45.00  & 56.87 & 87.57 & 5.41   & 14.79  & 25.63 & 54.53 \\
\rowcolor{gray!20} Liu~\cite{liu2019lending}                       & 100\%            & 40.79 & 66.82 & 76.36 & 96.12 & 46.96  & 68.28  & 75.48 & 92.01 & 19.21  & 35.97  & 43.30 & 60.69 \\
\rowcolor{gray!20} SAFA ~\cite{safa}                     & 100\%            & 81.15 & 94.23 & 96.85 & 99.49 & 78.28  & 91.60  & 93.79 & 98.15 & -      & -      & -     & -     \\
\rowcolor{gray!20} SAFA$^{\dag}$~\cite{safa}                    & 100\%            & 89.84 & 96.93 & 98.14 & 99.64 & 81.03  & 92.80  & 94.84 & 98.17 & -      & -      & -     & -     \\
\rowcolor{gray!20} DSM$^{\dag}$~\cite{shi2020looking}                    & 100\%            & 91.96 & 97.50 & 98.54 & 99.67 & 82.49  & 92.44  & 93.99 & 97.32 & -      & -      & -     & -     \\
\rowcolor{gray!20} L2LTR ~\cite{L2LTR}                    & 100\%            & 91.99 & 97.68 & 98.65 & 99.75 & 83.14  & 93.84  & 95.51 & 98.40 & 58.33  & 84.23  & 88.60 & 95.83 \\
GeoDTR~\cite{zhang2023cross}                    & 100\%            & 93.76 & 98.47 & 99.22 & 99.85 & 85.43  & 94.81  & 96.11 & 98.26 & 62.96  & 87.35  & 90.70 & 98.61 \\

TransGeo~\cite{zhu2022transgeo}                  & 100\%            & 94.08 & 98.36 & 99.04 & 99.77 & 84.95  & 94.14  & 95.78 & 98.37 & -      & -      & -     & -     \\

Sample4Geo$^{\ddag}$~\cite{Sample4Geo}                & 100\%            &97.83      &99.63      &99.75      & 99.89      &87.49      &96.56      & 97.50     & 98.98 & 60.57  & 89.50  & 92.99 & 98.92 \\ 
Ours-small                 & 100\%            &  93.53         & 98.42 &   99.18    & 99.77     & 84.44    & 94.85 & 96.15  &  98.53    &57.71        &86.35        &90.40       &98.49      \\ \hline
 Ours-small                 & 0\%          & 87.90 & 95.86 & 97.51  & 99.63      & 82.96       & 92.96 & 94.43      &  97.37     & 58.85       &84.27        &88.16       &97.45      \\ 
Ours-base                 & 0\%         & 92.56  & 97.67  & 98.55  & 99.61   &  84.58      &  93.95    & 95.29 & 97.59     &60.53        &86.35        &89.77       &97.52     \\ 
Ours-base$^{*}$                 & 10\%         & 94.88  & 98.80  & 99.36  & 99.77   &  87.89      &  95.27    & 96.40 & 98.37     &65.30        &89.47        &92.15       &98.27     \\
\hline \hline
\end{tabular}}
\caption{\textbf{Quantitative comparisons.} $\dag$ denotes satellite images are projected to ground view by polar transform~\cite{safa}.   $\ddag$ denotes results without hard negative sample sampling in ~\cite{Sample4Geo}. $*$ denotes we fine-tune unsupervised models on 10\% labeled images. The encoders of ConvNeXt-Small and ConvNeXt-Base are shortened as small and base~\cite{liu2022convnet}. \label{table:baseline}}
\vspace{-0.3cm}
\end{table*}

\subsection{Implementation Details} 
We train the CFP with the CycleGAN model~\cite{CycleGAN2017} for 50 epochs. We use ConvNeXt-Base~\cite{liu2022convnet} with 384 $\times$ 384 input images in Ours-base of \cref{table:baseline} and ConvNeXt-Small with 224 $\times$ 224 input images in other experiments on the CVUSA and CVACT datasets. We use ConvNeXt-Base~\cite{liu2022convnet} encoders with 384 $\times$ 768 like Sample4Geo~\cite{Sample4Geo} on VIGOR(Chicago). We use the AdamW~\cite{loshchilov2017decoupled} optimizers for 40 epochs of the cold-start stage, and 60 epochs of the semi-supervised stage.  The threshold of our filter in \cref{sec:semi-supervised} is 0.05 for CVUSA, 0.025 for CVACT, and 0.035 for VIGOR. Generally speaking, the higher threshold means the better quality of pseudo-labels (details in \cref{fig:pseudo-labels.pdf} (B)).

\subsection{Results of Unsupervised Learning}
As shown in \cref{table:baseline}, our unsupervised approach performs comparably to recent works on CVUSA, CVACT Val, and the large CVACT Test. This suggests that our model learns a robust retrieval and generalization ability without the supervision of human-annotated labels. When we fine-tune our unsupervised models on a few labeled images (10\%), the models perform better on CVACT Val and CVACT Test.  This further demonstrates the potential of our methods for unleashing unlabeled data.

\subsubsection{Analysis of Cold-Start Stage \label{sec:Cold-Start-ablation-study}}
In the cold-start stage, we aim to initialize a feature space where cross-view images from the same scene are aligned and thereby enable the models to predict some pairs as pseudo labels. So we remove some components to validate their functions by analyzing the R@1 evaluated in the validation sets and corresponding quality of pseudo labels in the training stage in \cref{table:stage-0}. 
\begin{table}[!htbp]
\resizebox{0.45\textwidth}{!}{
\centering

\begin{tabular}{c|ccc|ccc}
\hline \hline
\multirow{2}{*}{\begin{tabular}[c]{@{}c@{}}Ablation\\ (w/o)\end{tabular}} & \multicolumn{3}{c|}{CVUSA} & \multicolumn{3}{c}{CVACT} \\ \cline{2-7} 
                                                                          & R@1 $\uparrow$   & Labels  & Correct $\uparrow$ & R@1  $\uparrow$& Labels  & Correct $\uparrow$ \\ \hline
(a) BEV                                                                    &0.0225       &14         &0          &0.0      &6         &0          \\
(b) Fake                                                                   & 9.18      &119         &98          &22.78      &1349         &923          \\
(c) Intra                                                                  &15.58       &1476         &798          &42.98      &5539         &4281          \\
(d) Cross                                                                  &0.011       &21         &0          &0.011      &48         &0         \\ \hline
(e) Ours                                                                  &17.96       &1477         &968          &44.81      &6012         &4752    \\ \hline \hline     
\end{tabular}
}
\caption{\textbf{Ablation studies in the cold-start stage.} ``Labels" and ``Correct" denote the total and correct labels after the first stage.
\label{table:stage-0}}
\end{table}

\noindent \textbf{Projection:} In (a,b,e) of \cref{table:stage-0}, we show the effect of our correspondence-free projections on CVUSA and CVACT. When we remove the geometric projection (\ie, without BEV in (a)), the performance of our model drops significantly and fails to label a set, suggesting the model cannot align cross-view spatial information without prior geometry knowledge.  When we remove the imaging projections (\ie, without Fake in (b)), R@1 of the model also drops about half, because 
the different imaging representations of the same scene in different views hinder their alignments. Besides, the ground panoramas are not completed~\cite{wang2023finegrained} on CVUSA, which makes the geometric projection not work correctly and so the R@1 is lower than that on CVACT.

\noindent \textbf{Self-supervised contrastive learning:} In (c,e) of \cref{table:stage-0}, intra-view learning has been shown to enhance the quality of pseudo labels, primarily attributed to its ability to enhance scene discriminability and mitigate same-view feature overlap.  In (d) of \cref{table:stage-0}, without cross-view learning, the toward-zero performance suggests the importance of spatial correspondences again in CVGL.

\begin{figure*}[!t]
  \centering
  \centering
    \includegraphics[width=\linewidth]{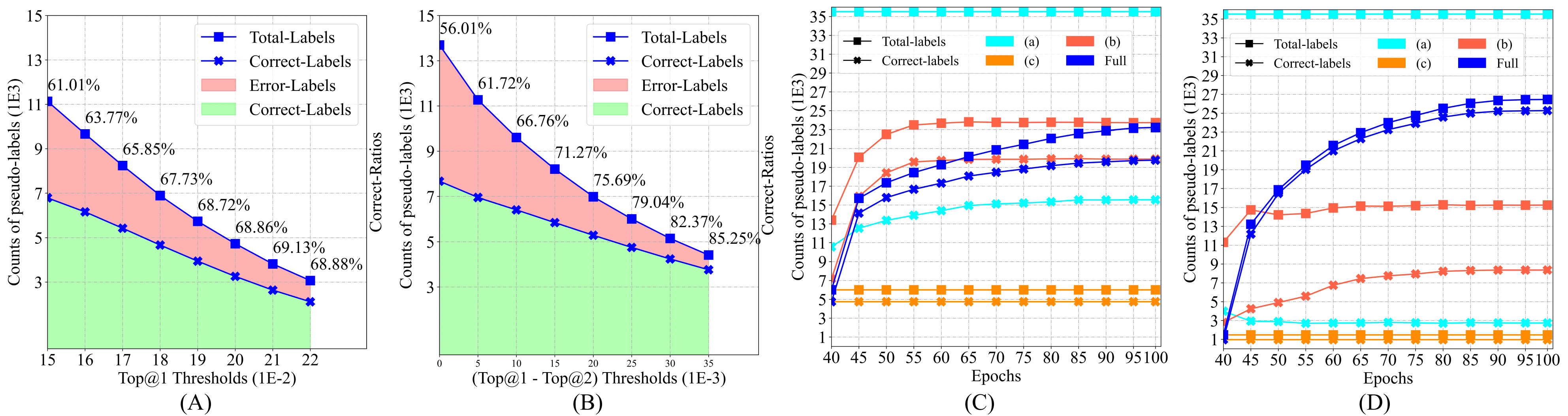}
    \vspace{-0.75cm}
    \caption{\textbf{Pseudo-labels.} The two left figures denote pseudo-labels produced by the highest or the difference value between the highest and second-highest retrieved similarity scores after the cold-start stage. The right two figures are the trend chart of pseudo-labels' counts of \cref{table:Stage-1} in (C) CVACT and (D) CVUSA. (a-c) denote our framework without \emph{mutual-matching}, \emph{threshold-filter}, and \emph{curriculum-learning}.}
  \label{fig:pseudo-labels.pdf}
  \vspace{-0.5cm}
\end{figure*}

\subsubsection{Analysis of Semi-Supervised Stage\label{sec:semi-supervised-ablation-study}}

In the semi-supervised stage of our study, our objective is to initiate robust training processes using pseudo-labeled ground-satellite image pairs and dynamically refine the pseudo-labels. Therefore, we delve into the design choices made during the assignment and refinement of pseudo labels, evaluating them based on the final performance in the validation sets and the changes observed in the correct labels during training.

\begin{table}[!htbp]
\centering
\resizebox{0.45\textwidth}{!}{

\begin{tabular}{c|cc|cc}
\hline \hline
\multirow{2}{*}{\begin{tabular}[c]{@{}c@{}}Ablation\\ (w/o)\end{tabular}} & \multicolumn{2}{c|}{CVUSA $\uparrow$} & \multicolumn{2}{c}{CVACT $\uparrow$} \\ \cline{2-5} 
                                                                          & R@1         & R@10         & R@1         & R@10        \\ \hline

%(a) linked to exp. xxx-s-224-ndm
(a) Mutual-Matching                                                        &  14.40  &   35.43           &   61.90          &     82.80       \\
(b) Threshold-Filter                                                       & 36.52            &   56.10           &    82.46         &        94.04      \\
(c) Curriculum-Learning                                                      &   32.56          &  54.35            &   63.14          &      82.83       \\ \hline 
Ours                                                                  &  87.90           &  97.51            &  82.96           &   94.43         \\ \hline \hline
\end{tabular}}
\caption{\textbf{Ablation studies in semi-supervised learning.} ``Full" denotes our Adaptive-Mutual-Matching.\label{table:Stage-1}
}
\end{table}
\noindent \textbf{Final performance:} In \cref{table:Stage-1}, we conduct some ablation studies to demonstrate the effectiveness of key technologies in the proposed adaptive mutual matching.

Firstly, we use the ground images and their most similar satellite images as initial pseudo-labels without our \emph{mutual-matching}, which gets poor performance, especially for CVUSA. This is because the proportion of error labels is too large in \cref{fig:pseudo-labels.pdf} (C-a) (about 70\% on CVACT), misleading the optimization direction. By digging into the unique characteristics of CVGL, we find that cross-view images are different representations of the same scene, so they should be successfully retrieved from each other. Therefore, we use the mutually matched cross-view images as pseudo-labels and improve their correct ratio from 11\% to 25\% on CVUSA and from 30\% to 53\% on CVACT.

Secondly, despite the improvement made by \emph{mutual-matching}, the correct ratio of assigned pseudo-labels is still low. With these pseudo-labels, CVACT has good performance because it has enough correct labels (about 7,000 correct and error pairs), but CVUSA has bad performance with more than 10,000 error labels and 2,000 correct labels. After filtering the labels by our \emph{threshold-filter} strategy, the correct ratio is improved from 25\% to 65\% in CVUSA and from 53\% to 80\% on CVACT (\eg, \cref{fig:pseudo-labels.pdf} (B)), and they both have a better final performance. So the \emph{threshold-filter} is very important for starting a robust retrieval system.

Lastly, we freeze the initial pseudo-labels, and the performance drops a lot because the model only learns limited knowledge based on fixed samples. So we introduce more hard samples by lowering the threshold of our \emph{threshold-filter} and matching more image pairs as the model's knowledge increases. After that, we get comparable performance compared to the recently supervised methods~\cite{Sample4Geo}.

\noindent \textbf{Labels change:} In the left two images of \cref{fig:pseudo-labels.pdf}, we compare two different filter choices. It's clear that filtering pseudo-labels by the maximum similarity scores has low accuracy, but our methods not only have a higher correct ratio but also allow the correct ratios to rise gradually as the threshold increases. In other words, we don't need to set a well-chosen threshold because the higher the threshold, the better the quality of pseudo-labels. Actually, the design of our \emph{threshold-filter} is motivated by a non-class characteristic like \cref{fig:pic/1/different_image_retrival.pdf}(C), each ground image should have the highest similarity with the corresponding satellite image and is dissimilar to images captured by other scenes.

In the right two images of \cref{fig:pseudo-labels.pdf}, it is observed that our \emph{mutual-matching} technology yields more precise initial pseudo-labels than direct predictions. The application of a \emph{threshold-filter} strategy further enhances the quality of these labels, while the incorporation of curriculum learning ideas ensures a gradual increase in the number of pseudo-labels with a consistent ratio of correctly labeled samples. Moreover, our \emph{threshold-filter} strategy is more important when initial labels are extremely noisy (\eg, CVUSA). Although it discards many pseudo-labels, the ratio of correct labels is guaranteed, which guides a robust and positively optimized direction for our model. Lastly, our \emph{curriculum-learning} increases the number of pseudo-labels with a stable ratio of correct labels step by step.

\subsection{Results of Semi-Supervised Learning \label{sec:semi-supervised-results}}
As shown in \cref{table:semi-supervised-learning-results}, we use a part of ground-truth labels to start semi-supervised learning on CVACT and VIGOR(Chicago).  Firstly, with the same labeled images, \ie, 10\% labeled images on CVACT and 30\% labeled images on VIGOR(Chicago), our method has better results than Sample4Geo~\cite{Sample4Geo}. In other words, we can improve the performance of existing supervised methods with our semi-supervised stage by unleashing unlabeled data. Secondly, we get good performance with only 355 images (1\%) on CVACT and 1274 images (10\%) on VIGOR(Chicago). This suggests our methods can work guided by a few labels and leverage unlabeled data to refine it for better performance. Lastly, the more ground-truth labels have a limited improvement on CVACT (\eg, from 5\% to 20\%), suggesting our method learns some common knowledge to automatically label most other images by leveraging a few samples. 
\begin{table}[!t]
\centering
\resizebox{0.45\textwidth}{!}{

\begin{tabular}{c|ccc|c|ccc}
\hline \hline
\multirow{2}{*}{\begin{tabular}[c]{@{}c@{}}GT\\ Ratio\end{tabular}} & \multicolumn{3}{c|}{CVACT} & \multirow{2}{*}{\begin{tabular}[c]{@{}c@{}}GT\\ Ratio\end{tabular}} & \multicolumn{3}{c}{VIGOR(Chicago)} \\ \cline{2-4} \cline{6-8} 
                                                                    & R@1     & R@5     & R@10   &                                                                     & R@1     & R@5    & R@10    \\ 
                                                                    \hline
\rowcolor{gray!50} 10\%                                                                & 56.10   & 81.69   & 88.18  & 30\%                                                                 &36.82         &65.16        &74.28          \\
\hline
1\%                                                                & 68.29   & 85.18   & 88.80  & 5\%                                                                 &25.82         &42.81        &49.81          \\
5\%                                                                & 78.10   & 90.87   & 93.11  & 10\%                                                                 &44.17         &63.30        &69.81         \\
10\%                                                                 & 78.88   & 91.31   & 93.53  & 20\%                                                                 &55.90         &75.44        &81.20         \\ 
20\%                                                                 & 79.60   & 91.98   & 93.96  & 30\%                                                                 &60.42         &80.12        &84.88         \\ \hline
100\%                                                                 & 84.44   & 94.85   & 98.53  & 100\%                                                                 & 68.40        & 88.49       &92.44        \\ \hline \hline
\end{tabular}}
\caption{\textbf{Results in semi-supervised settings.} The first gray line denotes the results of supervised Sample4Geo~\cite{Sample4Geo} without hard negative sample sampling and the others are our results. ``GT Ratio" denotes the ratio of ground-truth labels used for training.\label{table:semi-supervised-learning-results}
\vspace{-0.5cm}
}
\end{table}
\section{Discussion and Limitation}
The advanced supervised approach~\cite{Sample4Geo} discards spatial aggregation modules like SAFA~\cite{safa} and uses the share-weight encoders for training. But, without labels, discarding them brings some additional learning burdens which lead to suboptimal results in \cref{table:discussions} (a-b). Therefore, we still apply SAFA~\cite{safa} to help the model better understand and align the cross-view features and two non-share-weight encoders to improve the results. Besides, the entire training process is noisy, so the performance drops a lot without label smoothing in \cref{table:discussions} (c). 

In this work, we also utilize the localization and orientation alignment of ground and satellite images to simplify our unsupervised task like most supervised works\cite{Sample4Geo,zhu2022transgeo,safa}, but the future UCVGL should also get rid of these assumptions. So we also test the performance trained with random-shift or random-rotate images in \cref{table:discussions}, and we find our model is robust for them.

\begin{table}[!t]
\centering
\resizebox{0.43\textwidth}{!}{
% \usepackage{multirow}
\begin{tabular}{c|cccc}
\hline \hline
\multirow{2}{*}{Ablations} & \multicolumn{4}{c}{CVACT $\uparrow$} \\ \cline{2-5} 
                           & R@1  & R@5 & R@10 & R@1\% \\ \hline

(a) w/o safa                &  73.48    & 89.13    & 92.18     &  97.18     \\
(b) w/ share-weight        &  81.06    & 92.04    &  94.06    &    97.33   \\
(c) w/o label-smoothing     &  71.88    & 85.60    & 88.82     &  94.83    \\
(d) w/ random-shift           & 54.49     & 69.70    & 74.65     &  86.81     \\
(e) w/ random-rotate          &  50.32    & 68.50    & 75.08     &  88.70     \\
(f) Ours     &  82.96    & 92.96    & 94.43     &  97.37    \\

\hline \hline
\end{tabular}}
\caption{\textbf{Discussions of some common components.} 
\label{table:discussions}
\vspace{-0.3cm}}
\end{table}

While we make the unsupervised way work in CVGL, our method has some limitations. Firstly, to learn cross-view alignments without labels, we project ground panoramas to satellite-view by leveraging the prior knowledge of approximate spatial projection between cross-view images, but a fixed geometric projection is not enough for simulating various cross-view relationships in the real world. Secondly, our projection is homography like many works~\cite{safa,shi2022beyond,wang2023finegrained}, which is limited to represent the real transforms perfectly between two views only by images without extra 3D information (\eg, depth). For example, the advanced supervised cross-view image synthesis method~\cite{qian2023sat2density} does not work in cities of VIGOR because of the complex scene and serve occlusions.

\section{Conclusion}
In this paper, we propose the first framework to utilize unlabeled data in cross-view geo-localization. Compared to the recent supervised methods, our method achieves comparable performance in an unsupervised setting without any labels and in a semi-supervised setting with few labels.

\section{Acknowledgement} 
This work is supported by National Natural Science Foundation of China grants under contracts NO.62325111 and No.U22B2011. We thank the reviewers and ACs for their efforts. We want to thank the insightful discussions of Xiangyu Li, Jinwei Han, Jiakun Xu, Wuxuan Shi, Caoyuan Ma, Yajing Luo, Han Feng, Jiayu Li and Chao Pang. 

%%%%%%%%% REFERENCES
{\small
\bibliographystyle{ieee_fullname}
\bibliography{egbib}
}
\end{document}